\documentclass[11pt]{article}

\usepackage[preprint]{acl}
\makeatletter
\acl@anonymizefalse
\makeatother

\usepackage{times}
\usepackage{latexsym}
\usepackage[T1]{fontenc}
\usepackage[utf8]{inputenc}
\usepackage{microtype}
\usepackage{inconsolata}
\usepackage{graphicx}
\usepackage{booktabs}
\usepackage{multirow}
\usepackage{makecell}
\usepackage{tikz}
\usepackage{fontawesome5}
\usepackage{listings}
\usepackage{cleveref}

\lstdefinestyle{instructions}{
  basicstyle=\footnotesize\ttfamily,
  breaklines=true,
  columns=fullflexible,
  frame=single,
  numbers=left,
  numberstyle=\tiny,
  xleftmargin=2em,
  captionpos=b,
}
\lstdefinelanguage{json}{
  morestring=[b]",
  morekeywords={true,false,null},
  showstringspaces=false,
  literate=
   *{0}{{{\color[HTML]{7A3E9D}0}}}{1}
    {1}{{{\color[HTML]{7A3E9D}1}}}{1}
    {2}{{{\color[HTML]{7A3E9D}2}}}{1}
    {3}{{{\color[HTML]{7A3E9D}3}}}{1}
    {4}{{{\color[HTML]{7A3E9D}4}}}{1}
    {5}{{{\color[HTML]{7A3E9D}5}}}{1}
    {6}{{{\color[HTML]{7A3E9D}6}}}{1}
    {7}{{{\color[HTML]{7A3E9D}7}}}{1}
    {8}{{{\color[HTML]{7A3E9D}8}}}{1}
    {9}{{{\color[HTML]{7A3E9D}9}}}{1},
}
\lstdefinestyle{jsonschema}{
  style=instructions,
  language=json,
  basicstyle=\scriptsize\ttfamily,
  stringstyle=\color[HTML]{005A5A},
  keywordstyle=\color[HTML]{9A3412}\bfseries,
  numberstyle=\tiny\color[HTML]{666666},
  backgroundcolor=\color[HTML]{F8F8F8},
  rulecolor=\color[HTML]{CCCCCC},
}
\usetikzlibrary{arrows.meta,positioning,calc,shapes.geometric,shapes.misc,fit,backgrounds}

\newcommand{\system}{\texttt{schematize}}
\newcommand{\System}{\texttt{Schematize}}

\title{\System{}: An Agentic System for Generating and Refining\\
Information-Extraction Schemas for Legal Research}

  \author{Albert Sawczyn\textsuperscript{1}$^\dagger$ 
  \And  Jakub Binkowski\textsuperscript{1} 
  \And Kamil Tagowski\textsuperscript{1} 
  \AND  {\L}ukasz Augustyniak\textsuperscript{1} \And Berenika Kaczmarek-Templin\textsuperscript{1} \And Tomasz Kajdanowicz\textsuperscript{1} \AND
  \vspace{-1.5em} \\ \textsuperscript{1}Wroc{\l}aw University of Science and Technology
   \\ \vspace{-1em} \\ $^\dagger$albert.sawczyn@pwr.edu.pl}

\begin{document}
\maketitle

\begin{abstract}
Empirical legal research often relies on turning research questions into structured data extracted from large collections of rulings and
judgments. Designing the extraction schema and then extracting the data remain a manual, expertise-heavy bottleneck. We present \system{}, an open-source multi-agent system that interactively turns a researcher's problem statement into a validated extraction schema that can later be used for autonomous extraction. \System{} couples (i)~a clarification dialogue that elicits implicit expert intent, (ii)~iterative schema generation, (iii)~data-grounded
refinement that tests the schema against documents, and (iv)~chat-based post-editing. We evaluated the system with human legal professionals, introducing our novel methodology, and \system{} achieves top performance in most of tested configurations. While the system is designed to be domain-agnostic and applicable to any document collection, we tailor and evaluate it on legal research problems. We release \system{} as a pip-installable Python package with full documentation.
\end{abstract}

\section{Introduction}
\label{sec:introduction}

Empirical legal research often depends on structured representations of case law, rulings, and judgments: to study a question quantitatively, researchers must first extract fields that can be systematically analyzed across hundreds or thousands of documents \citep{hwang2022data,mali2024information}. Production legal-analytics platforms likewise combine large-scale document retrieval, expert-defined criteria, and structured extraction to support quantitative studies \citep{augustyniak2026bridging}. The quality of the outcomes depends strongly on how researchers formulate the research problem and then design the extraction schema that specifies \emph{what} to pull from each document. This upstream formulation is itself difficult: an initial request often leaves scope, entities, exclusions, comparisons, and downstream analytical goals implicit, even when the researcher has a clear study in mind. The process demands both legal expertise and data-modeling skills, from defining atomic fields to judging which values an LLM can reliably extract, while being slow, iterative, and prone to bias. For instance, a schema that omits a relevant field silently caps what the downstream study can ever discover.

Modern information-extraction systems share a common interface: they take a user-supplied \emph{schema} that declares what to pull from each document and ask a model to fill it in \citep{zaratiana2025gliner2,shrimal2025parse,langextract2026}, and the same assumption underlies legal IE pipelines \citep{cr2024legen,mali2024information}. A growing line of work instead tries to \emph{derive} schemas from a research question and a corpus \citep{levy2026schematiq,sadruddin2025llms4schemadiscovery,padmakumar2025intent}, but it typically validates against a reference schema and offers little support for grounding the schema in the documents it will be applied to. The upstream step -- turning a research question into a good schema and checking it against real data -- is thus left to the analyst, with no agreed way to tell a good schema from a bad one (Section~\ref{sec:related}).

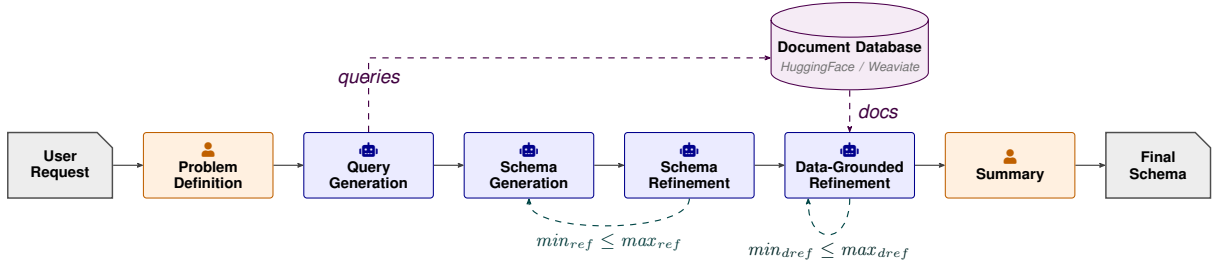
\begin{figure*}[t]
    \centering
    \begingroup
\newcommand{\subt}{\itshape\tiny\color{black!55}}
\newcommand{\humanicon}{{\color{orange!75!black}\faUser}}
\newcommand{\roboticon}{{\color{blue!55!black}\faRobot}}
\resizebox{\textwidth}{!}{%
\begin{tikzpicture}[
    font=\sffamily\scriptsize,
    >={Stealth[round,length=4pt]},
    node distance=5mm,
    block/.style={
        rounded corners=1.5pt, draw=blue!55!black, line width=0.5pt,
        fill=blue!8, text width=20mm, align=center,
        minimum height=11mm, inner sep=2.5pt},
    human/.style={block, draw=orange!75!black, fill=orange!12},
    io/.style={
        chamfered rectangle, chamfered rectangle corners=north east,
        chamfered rectangle xsep=4pt, draw=black!70, line width=0.7pt,
        fill=black!7, text width=13mm, align=center,
        minimum height=11mm, inner sep=2.5pt,
        font=\sffamily\scriptsize\bfseries},
    store/.style={
        draw=violet!60!black, line width=0.5pt, fill=violet!8,
        cylinder, shape border rotate=90, aspect=0.22,
        text width=25mm, align=center, minimum height=9mm, inner sep=2pt},
    sub/.style={font=\sffamily\footnotesize\itshape, text=black!55},
    desc/.style={font=\sffamily\scriptsize, text=black!70, align=left},
    flow/.style={->, line width=0.55pt, draw=black!75},
    loop/.style={->, line width=0.55pt, draw=teal!55!black, dashed},
    data/.style={->, line width=0.55pt, draw=violet!55!black, dashed},
]
\node[io] (start) {User\\Request};
\node[human, right=of start] (define)
    {\humanicon\\\textbf{Problem}\\\textbf{Definition}};
\node[block, right=of define] (query)
    {\roboticon\\\textbf{Query}\\\textbf{Generation}};
\node[block, right=of query] (gen)
    {\roboticon\\\textbf{Schema}\\\textbf{Generation}};
\node[block, right=of gen] (refine)
    {\roboticon\\\textbf{Schema}\\\textbf{Refinement}};
\node[block, right=of refine] (data)
    {\roboticon\\\textbf{Data-Grounded}\\\textbf{Refinement}};
\node[human, right=of data] (chat)
    {\humanicon\\\textbf{Summary}};
\node[io, right=of chat] (final) {Final\\Schema};

\node[store, above=7mm of data] (retriever)
    {\textbf{Document Database}\\[1pt]{\subt \mbox{HuggingFace} / \mbox{Weaviate}}};

\draw[flow] (start)  -- (define);
\draw[flow] (define) -- (query);
\draw[flow] (query)  -- (gen);
\draw[flow] (gen)    -- (refine);
\draw[flow] (refine) -- (data);
\draw[flow] (data)   -- (chat);
\draw[flow] (chat)   -- (final);

\draw[data] (query.north) |- (retriever.west) node[pos=0.25,above,sub,text=violet!50!black] {queries};
\draw[data] (retriever) -- (data) node[midway,right,sub,text=violet!50!black] {docs};

\draw[loop] (refine.south) .. controls +(0,-6mm) and +(0,-6mm) .. (gen.south)
    node[midway,below,sub,text=teal!45!black] {$\mathit{min}_{ref}\leq\mathit{max}_{ref}$};
\draw[loop] (data.south) .. controls +(0,-8mm) and +(0,-8mm) .. ([xshift=4mm]data.south west |- data.south)
    node[pos=0.5,below,sub,text=teal!30!black] {$\mathit{min}_{dref}\leq\mathit{max}_{dref}$};

\end{tikzpicture}
}
\endgroup
    \vspace{-1.7em}
    \caption{Overview of \system{}'s agentic pipeline. A user request is clarified, formalized into retrieval queries and an initial schema, refined via criteria-based (CBR, up to $\mathit{max}_{ref}$ iterations) and data-grounded (DGR, up to $\mathit{max}_{dref}$ iterations) loops, and finalized through a summary and interactive chat. \textcolor[HTML]{00008C}{Blue}~(\textcolor[HTML]{00008C}{\faRobot}) marks automated agent steps, \textcolor[HTML]{BF6000}{orange}~(\textcolor[HTML]{BF6000}{\faUser}) marks human-in-the-loop steps, \textcolor[HTML]{4D004D}{violet} marks the document store, and \textcolor[HTML]{004646}{teal} marks refinement loops.}
    \label{fig:architecture}
\end{figure*}

\System{} closes this gap with a human-in-the-loop, multi-agent pipeline
(Figure~\ref{fig:architecture}). A researcher states a problem in natural
language; a \emph{problem-definition-helper} agent asks clarifying
questions, and once the expert answers, a \emph{definer} agent turns the
request and responses into a formal problem definition.
The system then generates search
queries, drafts an initial schema, and improves it through two refinement
loops: a criteria-based refinement loop in which a \emph{critic} agent scores the schema against
quality criteria, and a data-grounded refinement loop that retrieves real documents with
the generated queries and tests whether the schema can actually be populated
from them. Finally, a summary explains how the schema was derived, and an interactive
chat lets the expert request further edits. The result is an extraction schema
that is grounded in both expert intent and real documents before any
large-scale extraction begins. Notably, we provide a novel methodology for the evaluation of schema quality by asking human experts to write exhaustive question sets for each research problem, which is specifically designed to limit the impact of biases in the evaluation process (\Cref{sec:evaluation}), and then scoring the schema for coverage by an LLM-as-judge \citep{10.5555/3666122.3668142_llmasjudge}.

\begin{itemize}
    \item \textbf{System \& library:} an open-source, pip-installable library for research-problem-to-schema generation that is model-agnostic -- any agent runs on either open- or closed-weight models -- with pluggable data connectors and a schema-based extractor.
    \item \textbf{Evaluation methodology \& data:} an expert-question coverage protocol -- experts write exhaustive question sets for each research problem ($\sim$15--30), and schemas are scored for coverage by an LLM-as-judge -- together with expert-curated legal research problems and their evaluation-questions sets.
    \item \textbf{Empirical study:} a comprehensive evaluation on legal research problems, comparing small and large LLMs, single- and multi-agent configurations, system ablations, and cost--quality trade-offs.
\end{itemize}

\paragraph{Availability.}
We publish \system{} as a pip-installable package\footnote{\url{https://pypi.org/project/schematize/}} with documentation\footnote{\url{https://pwr-ai.github.io/schematize/}} and source code\footnote{\url{https://github.com/pwr-ai/schematize}} available, all under the MIT license. The dataset containing all judgments used in the study is available in the Hugging Face repository.\footnote{\url{https://huggingface.co/datasets/JuDDGES/pl-court-raw} (CC BY 4.0 license)}

\section{Related Work}
\label{sec:related}

Information extraction (IE) turns unstructured documents into structured records \citep{JurafskyMartin2026}. As large language models have made extraction increasingly reliable \citep{Baietal2024}, the field has settled on a common interface: rather than hand-building an extractor for each task, modern systems take a user-supplied \emph{schema} that declares what to pull from each document and ask the model to fill it in \citep{zaratiana2025gliner2}.

\paragraph{Schema-driven information extraction} treats this schema as the contract between user and model. For instance, GLiNER2 unifies entity recognition, classification, and structured extraction behind a schema-based API \citep{zaratiana2025gliner2}; PARSE optimizes existing JSON schemas for reliable LLM extraction \citep{shrimal2025parse}; LangExtract grounds each  extracted field in its source span \citep{langextract2026}. The same assumption carries into legal IE, where systems extract entities and relations against task-specific schemas to support quantitative case-law analysis \citep{cr2024legen,hwang2022data,mali2024information}. 

\paragraph{Deriving schemas, rather than assuming them,} is the concern of the work closest to ours. \mbox{ScheMatiQ} \citep{levy2026schematiq} turns a research question and a corpus into a query-driven schema and creates a grounded database through interactive discovery; SchemaMiner builds schemas from scientific documents with human-in-the-loop refinement \citep{sadruddin2025llms4schemadiscovery}; intent-conditioned methods generate and refine literature-review table schemas \citep{padmakumar2025intent}; SciDaSynth assembles structured tables from multi-modal scientific sources \citep{wang2025scidasynth}; and LOGOS induces hierarchical schemas for qualitative grounded-theory analysis \citep{pi2025logos}. 

Our work differs from prior work along three axes. First, rather than assuming a corpus, \system{} generates retrieval queries from the problem statement and pulls documents through pluggable connectors, so schema design is decoupled from having a pre-assembled collection. Second, we separate a document-blind, criteria-based critique -- guarding against overfitting to any particular sample -- from a data-grounded refinement loop, whereas the systems above either skip document grounding or interleave it directly with generation. Third, and most importantly, we evaluate schemas against expert-authored \emph{question sets} rather than treating reference schemas as gold: closely related work often evaluates by reconstructing manual or table-derived schemas, even though such references can be ambiguous, shaped by feasibility constraints, or contain artifacts \citep{levy2026schematiq,padmakumar2025intent}. A question-coverage protocol is repeatable across system versions without re-annotation. Moreover, a good, robust schema for a given research problem can be constructed in multiple valid ways -- field names, granularity, and typing choices vary even among experts solving the same problem -- so schema design has no single ground-truth target; this is precisely why we evaluate against expert-authored questions rather than a single reference schema.

\paragraph{Empirical legal research} is our main focus, where automation must augment rather than replace expert judgment \citep{ramesh2025synthtexteval}. Prior legal IE work delivers strong extractors but, as above, leaves schema design to the analyst \citep{mali2024information,hwang2022data}. \System{} fills this gap with a human-in-the-loop pipeline that helps legal researchers express, ground obtained schema before any extraction begins.

\section{The \system{} System}
\label{sec:system}

\System{} is organized as a pipeline of LLM agents (Figure~\ref{fig:architecture}). Each agent has a single responsibility and communicates through structured messages. Users can configure the pipeline by choosing LLMs (either closed-weight API models or open-weight models), data connectors, and other parameters described in the library documentation. We describe each pipeline stage below. Notably, we tailored the prompts with the help of legal experts to be more specific to the legal domain.

\paragraph{Problem Definition}
First, the legal researcher sketches a problem of interest (e.g.,~``study personal-rights violations and assess their severity''). A \emph{problem-definition-helper} (PDH) agent reads this request and asks a short batch of clarifying questions -- about scope, jurisdiction, the granularity of the target variables, and intended downstream analysis -- surfacing implicit assumptions an expert could leave unstated. The expert answers, and a \emph{problem-definer} agent compiles the request and prepares a formal problem definition, including a problem statement, the legal domain, the scope of judgments of interest, the legal concepts involved, and a set of research questions.

\paragraph{Query Generation}
From the formal problem definition, a \emph{query-generator} agent produces search query to find the documents the schema will eventually be applied to. These queries feed the document retriever in the data-grounded loop (Section~\ref{sec:data-loop}).

\paragraph{Schema Generation and Criteria-Based Refinement}
A \emph{generator} agent first drafts an extraction schema from the problem definition; Listing~\ref{lst:schema-example} shows an excerpt. The schema then enters a criteria-based refinement (CBR) loop: an \emph{assessment} agent scores it for coverage of the problem definition, field clarity and non-redundancy, appropriate typing and granularity, and extractability, and a \emph{refiner} agent applies the recommended changes. To avoid overfitting the schema to the retrieved documents, the \emph{generator} and \emph{refiner} agents operate without document access and rely only on the problem-definition instructions.
The loop repeats until it runs at least $\mathit{min}_{ref}$ rounds and stops once the assessment reports no further changes are needed or a cap of $\mathit{max}_{ref}$ rounds is reached.

\paragraph{Data-Grounded Assessment and Refinement}
\label{sec:data-loop}
The second loop grounds the schema in the context of real documents. Using the generated queries, a document retriever fetches a set of matching legal documents through a configured backend, such as Hugging Face datasets or a Weaviate vector store. This data-grounded refinement (DGR) loop starts when a \emph{data-assessment} agent examines each retrieved document to assess the schema from a practical perspective, and a \emph{merger} agent consolidates these per-document assessments into a single set of revisions. Finally, a \emph{data-refiner} agent updates the schema accordingly. Like the first loop, it runs between $\mathit{min}_{dref}$ and $\mathit{max}_{dref}$ rounds, with the \emph{data-assessment} agent examining a configurable number of retrieved documents per round.

\paragraph{Summarization and Interactive Chat}
After the refinement loops, a \emph{summarizer} agent produces a short report explaining how the schema was derived -- which fields were added or changed in each loop and why -- so the expert can audit the process rather than trust an opaque output. The session then enters an open-ended chat: the expert can ask for further changes in natural language, and a chat agent edits the schema in place, keeping the human in control of the final artifact.

\subsection{Library Design}
\label{sec:library}
\System{} ships as an installable Python package so the pipeline can be run programmatically or embedded in other tools. LLM access goes through LiteLLM \citep{litellm}, and orchestrated through the langgraph framework \citep{langgraph2024}, making the library modular and model-agnostic: agents can be backed by either open-weight models (e.g.,~self-hosted Llama or Qwen run locally or behind a vLLM~\cite{kwon2023vllm}/Ollama~\cite{ollama} endpoint) or closed-weight API models (e.g.,~GPT, Claude, or Gemini). This lets users trade costs, latency, and data-privacy requirements against capability. Data access is abstracted through an abstract retriever base class, which ensures that the DGR loop receives documents in a uniform format regardless of their source. The library includes connectors for Hugging Face datasets and Weaviate, and supports other document sources through extensions provided by users.

\section{Empirical Evaluation}
\label{sec:evaluation}
We introduce a custom methodology for assessing the quality of generated schemas while limiting potential annotation biases. In particular, we evaluate the system on three Polish legal research problems in collaboration with seven legal experts.

\subsection{Methodology}
In prior studies, experts were asked to assess schemas generated by the evaluated systems. However, such holistic and intuitive evaluation can introduce confirmation and hindsight biases among annotators \citep{pmlr-v58-mahdavi17a}. Instead, we propose a protocol wherein domain experts cannot see the generated schema but independently write the atomic questions that the extracted data should be able to answer (see \Cref{fig:evaluation}). The experts wrote questions until they judged the topic to be exhausted; we suggested approximately 15--30 questions per problem.
By asking experts to formulate questions upfront, we shift their cognitive processing from post-hoc schema evaluation to direct reasoning about the problem, mitigating potential biases. For each research problem, experts wrote their questions from a single fixed PDH exchange: a user request, PDH clarifying questions from gpt-5.2, and the user's answers (\Cref{fig:evaluation}). In our study, we recruited 7 experts from the District Chamber of Legal Advisers in Wrocław and asked them to annotate three problems. We randomly assigned 3 experts to each problem. To ensure that experts understood the annotation task and were aligned on the protocol, we prepared detailed instructions (see \Cref{sec:appendix:evaluation}) and conducted a workshop session to explain the details and answer questions. 

\subsection{Model configuration and baselines} 
We compare \system{} with two baselines: a \emph{vanilla LLM}, which generates a schema in a single call, and the multi-agent \emph{ScheMatiQ} system \citep{levy2026schematiq}. Because \system{} includes the PDH stage, it receives information elicited from the user beyond the initial problem sketch. To isolate the contribution of the pipeline rather than this additional input, we evaluate each baseline both without and with the same fixed PDH exchange (\Cref{fig:evaluation}). This yields four baseline configurations: \emph{vanilla LLM}, \emph{vanilla LLM~+~PDH}, \emph{ScheMatiQ}, and \emph{ScheMatiQ~+~PDH}; the PDH-augmented configurations generate schemas from the clarified problem.
Our study uses both closed-weight models (gpt-5.4, gpt-5.4-mini, gpt-5.4-nano, claude-sonnet-4.6) and open-weight models (qwen3.6-35b-a3b, gemma-4-e4b-it, llama-4-scout-17b) of varying sizes\footnote{Licenses (all permit research use): gpt-5.4/-mini/-nano: OpenAI Services Agreement; claude-sonnet-4.6: Anthropic Commercial Terms; qwen3.6-35b-a3b: Apache 2.0; gemma-4-e4b-it: Apache 2.0; llama-4-scout-17b: Llama 4 Community License.}. In \Cref{sec:results}, we report coverage results for the best hyperparameter configuration found, and provide a broader hyperparameter sensitivity study in \Cref{sec:hyperparam} along with component ablations in \Cref{sec:ablation}.

\begin{table*}[!htb]
    \centering
    \small
    \resizebox{0.95\textwidth}{!}{%
    \setlength{\tabcolsep}{4.5pt}
    \begin{tabular}{lccccc}
        \toprule
        \textbf{Backbone} & \textbf{vanilla LLM} & \textbf{vanilla LLM + PDH} & \textbf{ScheMatiQ} & \textbf{ScheMatiQ + PDH} & \textbf{\system{}} \\
        \midrule
        gpt-5.4 & 36.3\% $\pm$ 11.7pp & \underline{64.3\% $\pm$ 19.7pp} & 43.8\% $\pm$ 15.6pp & 64.3\% $\pm$ 17.6pp & \textbf{79.4\% $\pm$ 12.6pp} \\
        gpt-5.4-mini & 31.7\% $\pm$ \phantom{0}7.2pp & 55.5\% $\pm$ 14.9pp & 23.9\% $\pm$ \phantom{0}7.9pp & \underline{57.6\% $\pm$ \phantom{0}7.6pp} & \textbf{74.8\% $\pm$ \phantom{0}7.2pp} \\
        gpt-5.4-nano & 22.2\% $\pm$ 18.0pp & \textbf{59.8\% $\pm$ 10.3pp} & 24.0\% $\pm$ 13.8pp & \underline{42.5\% $\pm$ 20.4pp} & 34.7\% $\pm$ 35.0pp \\
        claude-sonnet-4.6 & 37.1\% $\pm$ 10.8pp & 60.5\% $\pm$ 12.9pp & 49.6\% $\pm$ 29.2pp & \underline{67.1\% $\pm$ 14.8pp} & \textbf{70.0\% $\pm$ 18.5pp} \\
        qwen3.6-35b-a3b & 28.5\% $\pm$ \phantom{0}8.7pp & 51.6\% $\pm$ \phantom{0}2.5pp & 40.4\% $\pm$ 15.6pp & \textbf{67.2\% $\pm$ 15.3pp} & \underline{66.5\% $\pm$ 16.3pp} \\
        gemma-4-e4b-it & \phantom{0}3.8\% $\pm$ \phantom{0}4.0pp & 38.2\% $\pm$ 14.8pp & 34.5\% $\pm$ 18.2pp & \underline{46.5\% $\pm$ \phantom{0}4.8pp} & \textbf{53.4\% $\pm$ \phantom{0}3.7pp} \\
        llama-4-scout-17b & 27.7\% $\pm$ 20.0pp & 42.0\% $\pm$ \phantom{0}8.8pp & 32.7\% $\pm$ 19.3pp & \textbf{45.3\% $\pm$ 10.1pp} & \underline{45.1\% $\pm$ \phantom{0}7.9pp} \\
        \bottomrule
        \end{tabular}
    }
    \caption{Expert-question coverage (mean $\pm$ std, in pp) across backbones and configurations. ``+ PDH'' prepends the Problem-Definition-Helper stage to a baseline. Bold marks the best, underline the second-best, coverage per backbone.} 
    \label{tab:results}
\end{table*}

\subsection{Metrics} 
We report \textbf{coverage}: the fraction of expert questions that the generated schema can answer, judged per question by an LLM-as-judge (gpt-5.4-mini) that checks whether the schema contains the fields needed to resolve each question. LLM-as-a-judge evaluation is widely used and has demonstrated substantial alignment with human judgments in prior evaluation settings \citep{10.5555/3666122.3668142_llmasjudge,thakur2025judgingjudgesevaluatingalignment,janiak-etal-2025-illusion}. Thus, coverage provides a useful automatic proxy for schema quality.

\subsection{Results}
\label{sec:results}

\begin{figure*}[!htb]
  \centering
  \includegraphics[width=0.82\textwidth]{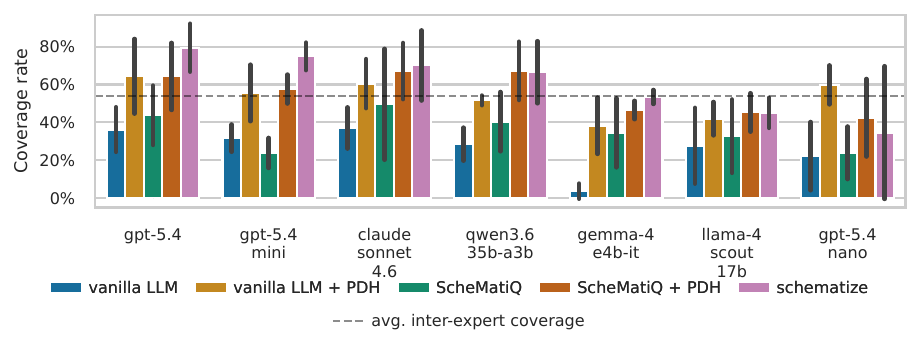}
  \vspace{-1.2em}
  \caption{Comparison of the \system{} to all tested baselines. The dashed line marks the average inter-expert coverage (\Cref{tab:expert_agreement}).}
  \label{fig:coverage_by_method_bars}
\end{figure*}

Table~\ref{tab:results} reports coverage across model sizes, with a complementary visualization in \Cref{fig:coverage_by_method_bars}, comparing \system{} with vanilla LLM and ScheMatiQ baselines. \System{} consistently outperforms both the vanilla LLM and ScheMatiQ baselines, as well as their PDH-augmented variants, across all tested backbones: it obtains the best or second-best result in 6 of the 7 tested configurations, including the best result in 4 configurations. This pattern holds for both open- and closed-weight models, suggesting that the proposed method remains effective in settings where privacy or deployment constraints favor open-weight models. Although larger models generally improve coverage, \system{} surpasses the average inter-expert coverage (\Cref{tab:expert_agreement}) on 4 of the 7 tested backbones. In other words, \system{} anticipates more of an expert's questions than the other experts do themselves, on average. On qwen3.6-35b-a3b and llama-4-scout-17b, ScheMatiQ~+~PDH edges out \system{} by only $\sim$1pp, and that margin comes from PDH -- our own contribution -- not from ScheMatiQ itself.

\begin{figure*}[!htb]
  \centering
  \includegraphics[width=0.88\textwidth]{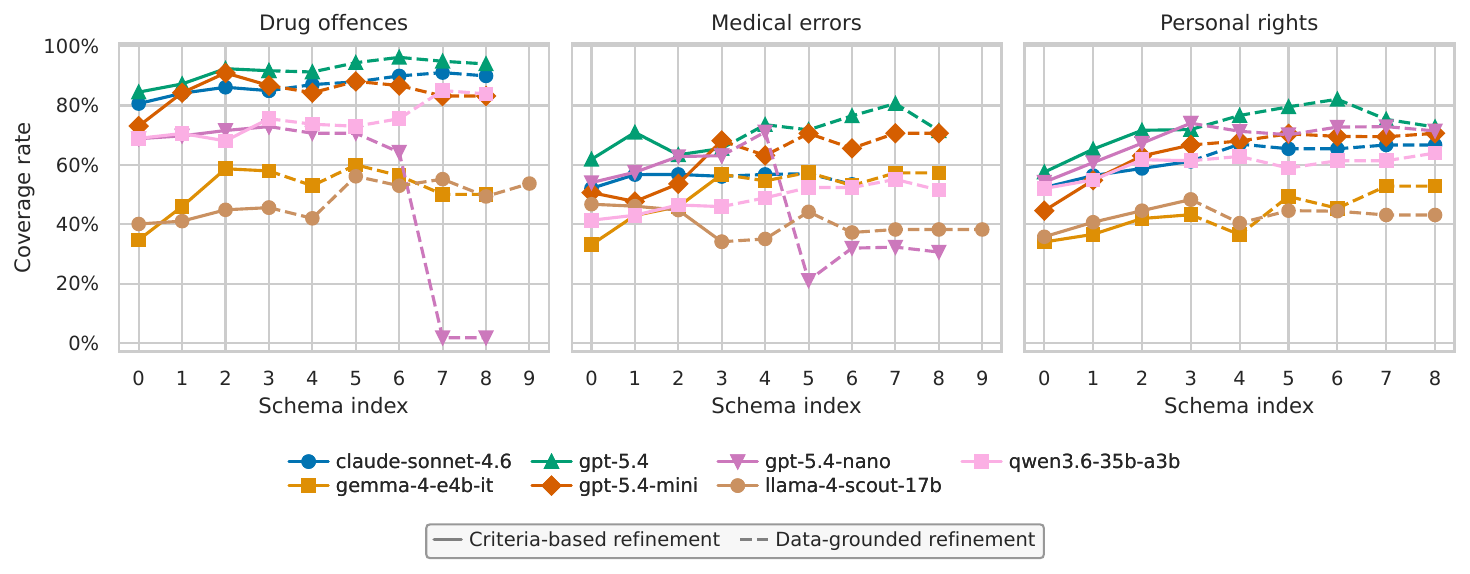}
  \vspace{-0.7em}
  \caption{Coverage change over subsequent iterations of schema refinements for three considered problems.}
  \label{fig:single_run_coverage_by_schema}
\end{figure*}

In addition, \Cref{fig:single_run_coverage_by_schema} shows that subsequent schema refinement iterations improve the coverage of the generated schema. However, these gains plateau after several iterations, suggesting that comparable coverage can be achieved without many expensive refinement steps. As \Cref{fig:single_run_coverage_by_schema} also shows, small models like gpt-5.4-nano struggle to grasp that their step is part of a larger pipeline: coverage drops sharply once the schema enters the data-grounded refinement (DGR) loop. On our inspection, the model appears to overfit to whichever document it is currently looking at, dropping previously established fields that are not reflected in that document instead of merging the new evidence with the existing schema. Further, we present detailed results in \Cref{sec:appendix:additional_results:detailed_results}, inter-expert coverage in \Cref{tab:expert_agreement}, and inference costs in \Cref{tab:costs}.

\section{Analysis}
\label{sec:analysis}

\subsection{Hyperparameter Sensitivity}
\label{sec:hyperparam}

Our system is parametrized by the minimum and maximum number of CBR rounds, $\mathit{min}_{ref}$ and $\mathit{max}_{ref}$, the minimum and maximum number of DGR rounds, $\mathit{min}_{dref}$ and $\mathit{max}_{dref}$, and the number of documents the \emph{data-assessment} agent examines per DGR round. To select the best hyperparameters, we run a grid search with gpt-5.4-nano and choose the configuration with the highest coverage.

\begin{figure*}[!htb]
    \centering
    \includegraphics[width=0.70\textwidth]{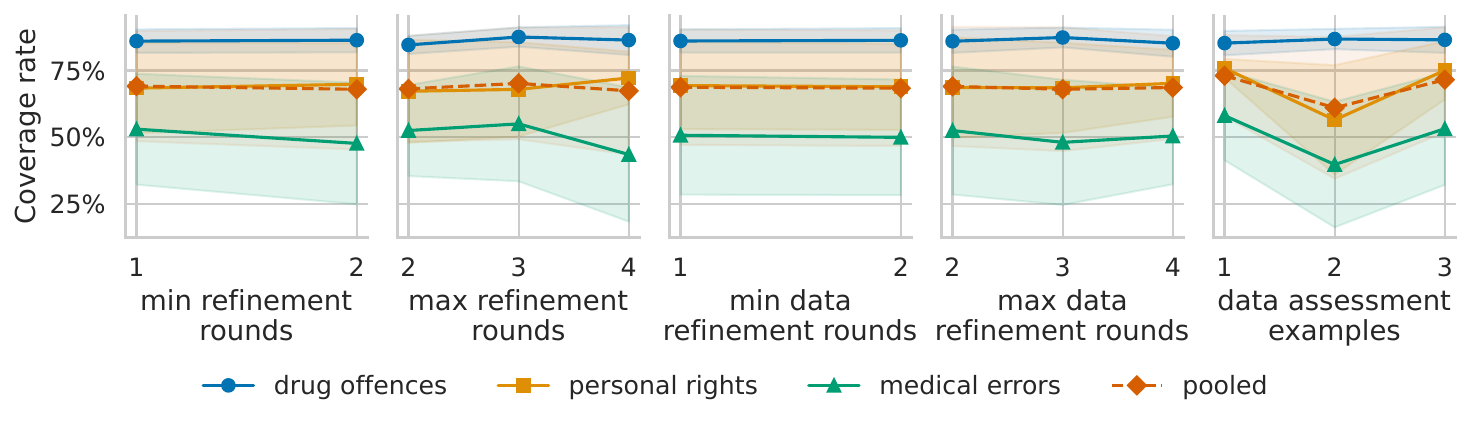}
   \vspace{-0.7em}
    \caption{Results of the hyperparameter sensitivity study. We measure coverage of the tested configurations for three legal research problems.}
    \label{fig:hyperparam}
\end{figure*}

We report coverage for the tested configurations in \Cref{fig:hyperparam}. Coverage remains stable across several hyperparameter settings, suggesting that \system{} is robust to these choices. This stability also indicates that users can often reduce costs by skipping hyperparameter search, especially when evaluation data is unavailable and the goal is schema generation. The best configuration, used for the main results in \Cref{tab:results}, sets $\mathit{min}_{ref}=2$, $\mathit{max}_{ref}=3$, $\mathit{min}_{dref}=2$, and $\mathit{max}_{dref}=4$, examining a single document per DGR round.

\subsection{Ablation Study}
\label{sec:ablation}
We assess the contribution of each pipeline stage by selectively ablating the problem-definition-helper (PDH), the criteria-based refinement (CBR) loop, and the data-grounded refinement (DGR) loop, individually and in combination; removing all three collapses \system{} into the single-LLM baseline (the vanilla LLM setting in \Cref{tab:results}). \Cref{tab:ablation} shows that removing CBR or DGR components has only a modest effect, well within noise, whereas removing PDH substantially decreases the performance. Likewise, removing two components together consistently leads to degradation, which indicates that they are complementary and \system{} peaks when PDH, CBR, and DGR operate together. The full system's larger variance reflects differences in problem difficulty rather than component instability. Ablation runs are separate from the main results, so the ``Full (\system{})'' row may differ slightly due to LLM non-determinism.

\section{Conclusion}
\label{sec:conclusion}

We contributed \system{}, a human-in-the-loop multi-agent system that turns a
research problem into a document-grounded extraction schema through clarification,
criteria-based critique, and data-grounded refinement against retrieved
documents. We rigorously tested the system by introducing a novel methodology to evaluate the schema quality with help of human legal professionals.
Empirical legal scholars and legal-tech builders can use \system{} to generate
extraction schemas from research problems without hand-crafting them first. The
code and pip-installable package are publicly available. 

\section{Limitations}
\label{sec:limitations}
Our evaluation is based on a single jurisdiction and language, and performance may vary for others. We focused on Polish case law because we were able to recruit legal experts with relevant expertise in this jurisdiction.

Coverage is scored by an LLM-as-judge, so the final estimate depends on the judge model and prompt. To limit biases that can arise when assesing a schema post-generation, experts authored question sets defining the target information needs prior to generation. However, we did not compare the LLM judge's individual coverage decisions with human annotations.

Additionally, coverage measures schema only recall quality: whether the generated fields cover expert information needs. While extra fields may hinder extraction of the fields of interest, they may also provide broader perspective and have only negligible effects on extraction quality. Also, our evaluation measures schema coverage rather than end-to-end extraction accuracy.

\section{Ethics Statement}
\System{} targets the legal domain, a high-stakes setting where automation must augment rather than replace expert judgment \citep{ramesh2025synthtexteval}. It is a research tool for designing extraction schemas and producing structured data for analysis; it does not provide legal advice, and its outputs should not be relied on for legal decisions. Because automatically generated schemas can look authoritative, we keep the human in the loop at both problem definition and final editing, and the summarizer makes the derivation auditable to discourage over-reliance. 

The study with human experts was carried out in collaboration with the District Chamber of Legal Advisers in Wroc{\l}aw, Poland, under a formal agreement between the Chamber and Wroc{\l}aw University of Science and Technology. The experts participated voluntarily and without compensation, with the exception of the study coordinator, who is an employee funded by the National Science Centre grant.

\section*{Acknowledgments}
\paragraph*{Annotators} We gratefully thank the legal experts from the District Chamber of Legal Advisers in Wroc{\l}aw, Poland, who contributed the annotations: Monika Boche{\'n}ska, Dominika Fikus, Dorota Jarz\k{e}bowska, Berenika Kaczmarek-Templin, Ma{\l}gorzata Koz{\l}owska, Maciej Kusyk, and Mateusz Lato.
Their domain expertise was essential for constructing the evaluation protocol and expert question sets.
\paragraph*{Funding} This work was co-funded by the National Science Centre, Poland, under CHIST-ERA Open \& Re-usable Research Data \& Software (grant no. 2022/04/Y/ST6/00183). We gratefully acknowledge the Wroclaw Center for Networking and Supercomputing for providing computing facilities and support.

\bibliography{references}

\clearpage

\appendix

\section{Example Schemas}
\label{app:schemas}

Listing~\ref{lst:schema-example} shows an excerpt from one generated
schema. Due to space constraints, we present only an illustrative portion of the full schema.

\noindent\begin{minipage}{\linewidth}
\begin{lstlisting}[style=jsonschema,basicstyle=\tiny\ttfamily,caption={Excerpt of a generated extraction schema.},label={lst:schema-example}]
{
  "fields": [
    {
      "name": "case_id",
      "type_": "string",
      "description": "Unique judgment or record identifier."
    },
    {
      "name": "judgment_year",
      "type_": "integer",
      "description": "Year when the judgment was issued."
    },
    {
      "name": "outcome",
      "type_": "enum",
      "enum_values": [
        "skazanie_bez_zawieszenia",
        "skazanie_z_zawieszeniem",
        "warunkowe_umorzenie",
        "inne_nieustalone"
      ],
      "description": "Type of judgment outcome."
    },
    {
      "name": "age_years",
      "type_": "integer",
      "description": "Age in years for the primary defendant."
    },
    ...
  ]
}
\end{lstlisting}
\end{minipage}

\section{Expert-Based Evaluation Protocol}
\label{sec:appendix:evaluation}

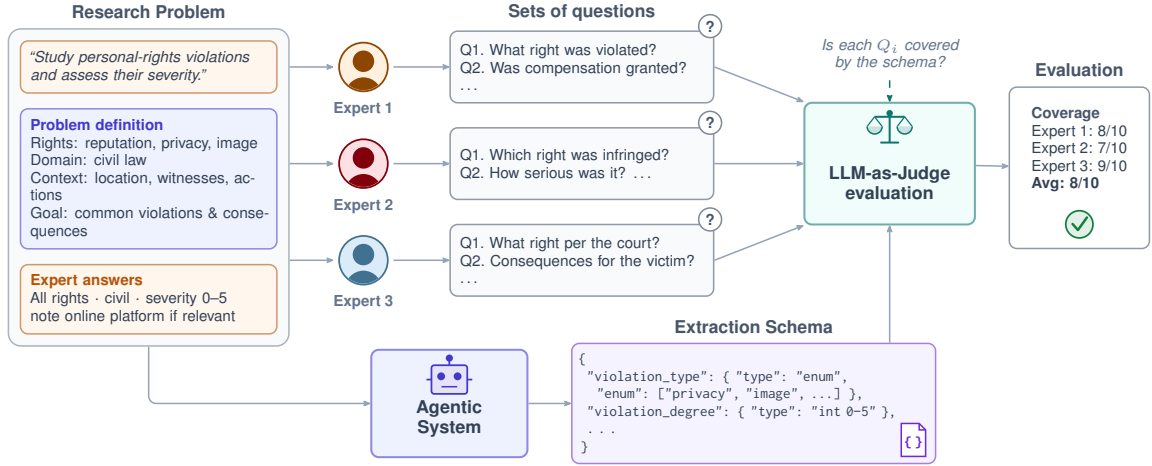
\begin{figure*}[t]
    \centering
    \begingroup
\definecolor{figInk}{HTML}{334155}      
\definecolor{figMut}{HTML}{64748B}      
\definecolor{figLine}{HTML}{94A3B8}     
\definecolor{figAmber}{HTML}{B45309}    
\definecolor{figAmberBg}{HTML}{FFF7ED}
\definecolor{figIndigo}{HTML}{4338CA}   
\definecolor{figIndigoBg}{HTML}{EEF2FF}
\definecolor{figViolet}{HTML}{6D28D9}   
\definecolor{figVioletBg}{HTML}{F5F3FF}
\definecolor{figTeal}{HTML}{0F766E}     
\definecolor{figTealBg}{HTML}{F0FDFA}
\definecolor{figSlate}{HTML}{475569}    
\definecolor{figSlateBg}{HTML}{F8FAFC}
\definecolor{figGreen}{HTML}{15803D}
\resizebox{0.95\textwidth}{!}{%
\begin{tikzpicture}[
    font=\sffamily\scriptsize, text=figInk,
    >={Stealth[round,length=4pt]},
    title/.style={font=\sffamily\scriptsize\bfseries, text=figSlate},
    card/.style 2 args={
        rounded corners=3pt, line width=0.6pt, inner sep=4pt,
        text width=#1, align=left, font=\sffamily\tiny,
        draw=#2!60!white, fill=white},
    human/.style={card={33mm}{figAmber}, fill=figAmberBg},
    auto/.style={card={33mm}{figIndigo}, fill=figIndigoBg},
    qset/.style={card={34mm}{figSlate}, fill=figSlateBg,
        minimum height=10mm, anchor=west},
    schema/.style={card={48mm}{figViolet}, fill=figVioletBg,
        font=\ttfamily\tiny, anchor=west, minimum height=13mm},
    agent/.style={card={19mm}{figIndigo}, fill=figIndigoBg, align=center,
        minimum height=15mm, font=\sffamily\scriptsize\bfseries,
        line width=0.8pt},
    judge/.style={card={21mm}{figTeal}, fill=figTealBg, align=center,
        minimum height=17mm, font=\sffamily\scriptsize\bfseries,
        line width=0.8pt},
    elab/.style={font=\sffamily\tiny\bfseries, text=figMut},
    sub/.style={font=\sffamily\tiny\itshape, text=figMut, align=center},
    flow/.style={->, line width=0.6pt, draw=figLine, rounded corners=3pt},
    avatar/.pic={
        \draw[#1!55!black, line width=0.7pt, fill=#1!12!white]
            (0,0) circle (3.4mm);
        \begin{scope}
            \clip (0,0) circle (3.25mm);
            \fill[#1!55!black] (0,-4.35mm) ellipse [x radius=2.7mm, y radius=2.9mm];
            \fill[#1!55!black] (0,0.8mm) circle (1.5mm);
        \end{scope}},
    robot/.pic={
        \draw[figIndigo, line width=0.7pt] (0,2.1mm) -- (0,2.9mm);
        \fill[figIndigo] (0,3.1mm) circle (0.5mm);
        \draw[figIndigo, line width=0.7pt, fill=figIndigo!14!white,
              rounded corners=1.1pt]
            (-2.5mm,-1.9mm) rectangle (2.5mm,2.1mm);
        \fill[figIndigo] (-1.15mm,0.55mm) circle (0.55mm);
        \fill[figIndigo] (1.15mm,0.55mm) circle (0.55mm);
        \draw[figIndigo, line width=0.6pt, rounded corners=1pt]
            (-1.1mm,-0.9mm) -- (1.1mm,-0.9mm);
        \draw[figIndigo, line width=0.7pt] (-2.5mm,0.1mm) -- (-3.2mm,0.1mm)
            (2.5mm,0.1mm) -- (3.2mm,0.1mm);},
    scales/.pic={
        \draw[figTeal, line width=0.8pt] (0,-2.1mm) -- (0,2.0mm);
        \fill[figTeal] (0,2.0mm) circle (0.45mm);
        \draw[figTeal, line width=0.8pt] (-2.1mm,1.5mm) -- (2.1mm,1.5mm);
        \draw[figTeal, line width=0.8pt, line cap=round]
            (-1.2mm,-2.1mm) -- (1.2mm,-2.1mm);
        \foreach \s in {-1,1}{
            \draw[figTeal, line width=0.5pt]
                (\s*2.1mm,1.5mm) -- ++(-1.05mm,-1.35mm)
                (\s*2.1mm,1.5mm) -- ++(1.05mm,-1.35mm);
            \draw[figTeal, line width=0.6pt, fill=figTeal!22!white]
                (\s*2.1mm,0.15mm) ++(-1.05mm,0)
                arc[start angle=180, end angle=360, radius=1.05mm];}},
    doc/.pic={
        \draw[figViolet, line width=0.6pt, fill=white]
            (-1.7mm,-2.3mm) -- (-1.7mm,2.3mm) -- (0.8mm,2.3mm)
            -- (1.7mm,1.4mm) -- (1.7mm,-2.3mm) -- cycle;
        \draw[figViolet, line width=0.6pt] (0.8mm,2.3mm)
            -- (0.8mm,1.4mm) -- (1.7mm,1.4mm);
        \node[font=\ttfamily\tiny\bfseries, text=figViolet] at (0,-0.35mm)
            {\{\,\}};},
    check/.pic={
        \draw[figGreen, line width=0.7pt, fill=figGreen!12!white]
            (0,0) circle (1.9mm);
        \draw[figGreen, line width=1pt, line cap=round, line join=round]
            (-0.95mm,0.05mm) -- (-0.2mm,-0.8mm) -- (1.05mm,0.95mm);},
    qmark/.pic={
        \draw[figSlate!70!white, line width=0.6pt, fill=white]
            (0,0) circle (1.6mm);
        \node[font=\sffamily\tiny\bfseries, text=figSlate] at (0,0) {?};},
]

\node[human] (req) at (0,0)
    {\textit{``Study personal-rights violations and assess their severity.''}};
\node[auto, below=2mm of req.south west, anchor=north west] (pdef) {%
    \textbf{\textcolor{figIndigo}{Problem definition}}\\
    Rights: reputation, privacy, image\\
    Domain: civil law\\
    Context: location, witnesses, actions\\
    Goal: common violations \& consequences};
\node[human, below=2mm of pdef.south west, anchor=north west] (ans) {%
    \textbf{\textcolor{figAmber}{Expert answers}}\\
    All rights $\cdot$ civil $\cdot$ severity 0--5\\
    note online platform if relevant};
\begin{scope}[on background layer]
    \node[draw=figSlate!45!white, rounded corners=5pt, line width=0.6pt,
          fill=figSlateBg, fit=(req)(pdef)(ans), inner sep=4.5pt] (rp) {};
\end{scope}
\node[title, anchor=south] at (rp.north) {Research Problem};

\coordinate (e1) at (3.0, 0.0);
\coordinate (e2) at (3.0,-1.35);
\coordinate (e3) at (3.0,-2.7);
\pic at (e1) {avatar=orange};
\pic at (e2) {avatar=red!70!purple};
\pic at (e3) {avatar=cyan!70!blue};
\node[elab, below=3.6mm of e1] {Expert 1};
\node[elab, below=3.6mm of e2] {Expert 2};
\node[elab, below=3.6mm of e3] {Expert 3};

\node[qset] (q1) at (4.2, 0.0)
    {Q1.\ What right was violated?\\ Q2.\ Was compensation granted? \dots};
\node[qset] (q2) at (4.2,-1.35)
    {Q1.\ Which right was infringed?\\ Q2.\ How serious was it? \dots};
\node[qset] (q3) at (4.2,-2.7)
    {Q1.\ What right per the court?\\ Q2.\ Consequences for the victim? \dots};
\foreach \i in {1,2,3}{ \pic at ([xshift=-0.6mm,yshift=0.6mm]q\i.north east) {qmark}; }
\node[title, anchor=south] at (q1.north) {Sets of questions};

\node[agent] (agent) at (4.2,-4.7) {\vspace{3.2mm}\\Agentic\\System};
\pic at ([yshift=-4.4mm]agent.north) {robot};
\node[schema] (schema) at (5.9,-4.7)
    {\{\\
    \ "violation\_type": \{ "type": "enum",\\
    \ \ \ "enum": ["privacy", "image", ...] \},\\
    \ "violation\_degree": \{ "type": "int 0--5" \},\\
    \ \dots\\
    \}};
\pic at ([xshift=-3.2mm,yshift=3.4mm]schema.south east) {doc};
\node[title, anchor=south] at (schema.north) {Extraction Schema};

\node[judge] (judge) at (10.35,-1.35) {\vspace{4.2mm}\\LLM-as-Judge\\ evaluation};
\pic at ([yshift=-3.6mm]judge.north) {scales};
\node[sub, above=3mm of judge, text width=30mm] (jq)
    {Is each $Q_i$ covered\\ by the schema?};
\node[font=\sffamily\tiny, text=figInk, align=left] (outtxt) at (13.0,-1.15)
    {\textbf{\textcolor{figSlate}{Coverage}}\\
    Expert 1: 8/10\\
    Expert 2: 7/10\\
    Expert 3: 9/10\\
    \textbf{Avg: 8/10}};
\node[circle, minimum size=3.8mm] (checkmark) at ([yshift=-3.2mm]outtxt.south) {};
\pic at (checkmark) {check};
\begin{scope}[on background layer]
    \node[draw=figSlate!60!white, rounded corners=3pt, line width=0.6pt,
          fill=white, fit=(outtxt)(checkmark), inner xsep=5pt, inner ysep=4pt]
        (out) {};
\end{scope}
\node[title, anchor=south] at (out.north) {Evaluation};

\foreach \i in {1,2,3}{ \draw[flow] (rp.east |- e\i) -- ([xshift=-4.2mm]e\i); }
\draw[flow] ([xshift=4.2mm]e1) -- (q1.west);
\draw[flow] ([xshift=4.2mm]e2) -- (q2.west);
\draw[flow] ([xshift=4.2mm]e3) -- (q3.west);

\draw[flow] (rp.south) |- (agent.west);
\draw[flow] (agent.east) -- (schema.west);

\draw[flow] (q1.east) -- (judge.north west);
\draw[flow] (q2.east) -- (judge.west);
\draw[flow] (q3.east) -- (judge.south west);
\draw[flow] (judge.south |- schema.north) -- (judge.south);
\draw[flow, dashed, draw=figTeal] (jq) -- (judge);
\draw[flow] (judge) -- (out);

\end{tikzpicture}
}
\endgroup
    \caption{Expert-based evaluation pipeline. Given a shared research
    problem, each domain expert writes an
    exhaustive question set the extracted data should answer, while the
    \textcolor[HTML]{00008C}{agentic system} independently produces an
    extraction \textcolor[HTML]{4D004D}{schema}. An
    \textcolor[HTML]{004646}{LLM-as-judge} checks whether each expert
    question is covered by the schema, yielding per-expert and average
    coverage scores.}
    \label{fig:evaluation}
\end{figure*}

\subsection{Expert Instructions}
\label{sec:appendix:evaluation:instructions}
Listing~\ref{lst:expert-instructions} shows a shortened version of the annotation instructions given to the experts (the detailed instructions referenced in Section~\ref{sec:evaluation}).

\subsection{Inter-expert Coverage}
\label{sec:appendix:evaluation:agreement}
\begin{table}[!htb]
    \centering
    \small
    \setlength{\tabcolsep}{4.5pt}
    \begin{tabular}{lcc}
        \toprule
        \textbf{Problem} & \textbf{Questions/expert} & \textbf{Inter-expert coverage} \\
        \midrule
        Drug offences & 28.3 $\pm$ \phantom{0}9.6 & 56.4\% $\pm$ \phantom{0}7.9pp \\
        Medical errors & 33.0 $\pm$ 13.0 & 52.7\% $\pm$ 19.0pp \\
        Personal rights & 26.0 $\pm$ \phantom{0}3.0 & 52.3\% $\pm$ 17.1pp \\
        Overall & 29.1 $\pm$ \phantom{0}8.8 & 53.8\% $\pm$ 14.6pp \\
        \bottomrule
        \end{tabular}
    \caption{Average number of questions written per expert and inter-expert coverage (\Cref{sec:appendix:evaluation:agreement}) per research problem.}
    \label{tab:expert_agreement}
\end{table}

Since each research problem was annotated by three experts independently, we can measure how much their question sets overlap. We compute \emph{inter-expert coverage} with the same LLM-as-judge protocol used for schema coverage (\Cref{sec:evaluation}), but applied between experts: for each ordered pair of experts $(a, b)$ on the same problem, we check how many of expert $a$'s questions are covered by expert $b$'s question set alone. Since this is not symmetric, we compute it for both $(a, b)$ and $(b, a)$ for every pair, then average over all ordered pairs and problems. \Cref{tab:expert_agreement} reports a mean ordered-pair inter-expert coverage of 53.8\%, underscoring that formulating a complete extraction schema is inherently difficult and admits no single ground truth. \System{} exceeds this average benchmark on several backbones (\Cref{sec:results}), meaning that its schemas cover more benchmark questions than are covered, on average, by another expert's question set; this does not imply superiority to every individual expert.

\noindent\begin{minipage}{\linewidth}
  \begin{lstlisting}[style=instructions,numbers=none,basicstyle=\tiny\ttfamily,caption={Shortened version of the expert annotation instructions.},label={lst:expert-instructions}]
  Annotator Instructions Summary
  Role and Task: 
  The annotator creates a set of questions that a data extraction schema must answer based on Polish court judgments. These questions act as a benchmark to verify if the automated schema captures all key information defined by the expert.
  Research Problem Definition:
  The problem is defined via a three-part dialogue:
      1.	user_input: The researcher describes the research goal.
      2.	problem_help: The bot asks clarifying questions.
      3.	user_feedback: The researcher confirms, rejects, or adds details.
  Questions must stem from the entire dialogue, covering only the confirmed scope and ignoring rejected topics.
  Criteria for Valid Questions:
  Questions must determine what the system can extract for quantitative analysis.
      1.	Extractability: Questions must be answerable directly from a single judgment's text, without external knowledge or legal doctrine.
      2.	Unambiguity: Questions must be clear in intent. Subjective evaluations are permitted if grounded in the text.
      3.	Simplicity: One question equals one fact. Multi-threaded questions must be split.
      4.	Coverage: The questions must comprehensively cover all aspects confirmed in the dialogue.
  Answer Formats:
  Acceptable formats include Yes/No, Number, Scale, Text, and Category. If categorical options are not exhaustive, use a series of Yes/No questions instead.
  \end{lstlisting}
  \end{minipage}

\section{Detailed results}
\label{sec:appendix:additional_results:detailed_results}

\begin{table*}[!htb]
    \centering
    \resizebox{\textwidth}{!}{%
    \begin{tabular}{l|ccccc|ccccc|ccccc}
        \toprule
         & \multicolumn{5}{c}{\textbf{Drug offences}} & \multicolumn{5}{c}{\textbf{Medical errors}} & \multicolumn{5}{c}{\textbf{Personal rights}} \\
        \textbf{Backbone} & \makecell{\textbf{vanilla}\\\textbf{LLM}} & \makecell{\textbf{vanilla}\\\textbf{LLM}\\\textbf{+ PDH}} & \makecell{\textbf{Sche}\\\textbf{MatiQ}} & \makecell{\textbf{Sche}\\\textbf{MatiQ}\\\textbf{+ PDH}} & \textbf{\system{}} & \makecell{\textbf{vanilla}\\\textbf{LLM}} & \makecell{\textbf{vanilla}\\\textbf{LLM}\\\textbf{+ PDH}} & \makecell{\textbf{Sche}\\\textbf{MatiQ}} & \makecell{\textbf{Sche}\\\textbf{MatiQ}\\\textbf{+ PDH}} & \textbf{\system{}} & \makecell{\textbf{vanilla}\\\textbf{LLM}} & \makecell{\textbf{vanilla}\\\textbf{LLM}\\\textbf{+ PDH}} & \makecell{\textbf{Sche}\\\textbf{MatiQ}} & \makecell{\textbf{Sche}\\\textbf{MatiQ}\\\textbf{+ PDH}} & \textbf{\system{}} \\
        \midrule
        gpt-5.4 & 49.0\% & \underline{87.0\%} & 59.3\% & 84.3\% & \textbf{94.0\%} & 25.9\% & 53.4\% & 28.1\% & \underline{57.2\%} & \textbf{71.5\%} & 33.9\% & \underline{52.5\%} & 43.9\% & 51.2\% & \textbf{72.8\%} \\
        gpt-5.4-mini & 38.0\% & \underline{72.7\%} & 15.9\% & 66.3\% & \textbf{83.2\%} & 23.9\% & 47.9\% & 31.7\% & \underline{54.1\%} & \textbf{70.6\%} & 33.3\% & 46.0\% & 24.1\% & \underline{52.4\%} & \textbf{70.6\%} \\
        gpt-5.4-nano & 41.2\% & \textbf{71.5\%} & \phantom{0}9.2\% & \underline{63.6\%} & \phantom{0}1.9\% & 20.0\% & \textbf{52.3\%} & 26.1\% & 22.8\% & \underline{30.7\%} & \phantom{0}5.5\% & \underline{55.6\%} & 36.6\% & 41.0\% & \textbf{71.4\%} \\
        claude-sonnet-4.6 & 49.5\% & 74.9\% & 82.6\% & \underline{84.1\%} & \textbf{89.9\%} & 29.7\% & 50.1\% & 27.3\% & \textbf{57.2\%} & \underline{53.5\%} & 32.1\% & 56.4\% & 38.8\% & \underline{59.9\%} & \textbf{66.7\%} \\
        qwen3.6-35b-a3b & 27.9\% & 54.5\% & 47.3\% & \textbf{84.0\%} & \underline{83.9\%} & 20.2\% & 50.1\% & 22.6\% & \textbf{54.0\%} & \underline{51.5\%} & 37.5\% & 50.2\% & 51.4\% & \underline{63.5\%} & \textbf{64.0\%} \\
        gemma-4-e4b-it & \phantom{0}0.0\% & \textbf{55.0\%} & \underline{53.0\%} & 52.0\% & 50.1\% & \phantom{0}3.4\% & 26.9\% & 16.6\% & \underline{43.2\%} & \textbf{57.3\%} & \phantom{0}7.9\% & 32.8\% & 33.9\% & \underline{44.4\%} & \textbf{52.9\%} \\
        llama-4-scout-17b & 25.0\% & 50.1\% & \underline{54.2\%} & \textbf{55.8\%} & 53.7\% & \phantom{0}9.2\% & 32.7\% & 16.8\% & \underline{35.6\%} & \textbf{38.3\%} & \textbf{48.8\%} & 43.2\% & 27.3\% & \underline{44.4\%} & 43.2\% \\
        \bottomrule
        \end{tabular}
    }
    \caption{Per-problem expert-question coverage (\%) across backbones and configurations; columns match \Cref{tab:results}. Bold marks the best, underline the second-best, configuration per backbone and problem.}
    \label{tab:results-detailed}
\end{table*}

While \Cref{tab:results} averages coverage across the three research problems, \Cref{tab:results-detailed} reveals substantial variation across both problems and backbones. Averaged across backbones, ScheMatiQ~+~PDH performs best on drug offences (70.0\%), whereas \system{} performs best on medical errors (53.3\%) and personal rights (63.1\%). Across the 21 backbone-problem combinations, \system{} ranks first in 13 and first or second in 17. Its performance is particularly consistent on personal rights, where it leads for six of the seven backbones, and on medical errors, where it ranks among the top two for every backbone. Results on drug offences are more mixed: although \system{} wins for three backbones, ScheMatiQ~+~PDH achieves the highest cross-backbone average.

\subsection{Ablation}
\label{sec:appendix:additional_results:ablation}

\begin{table}[!htb]
    \centering
    \footnotesize
    \setlength{\tabcolsep}{3pt}
    \resizebox{\columnwidth}{!}{%
        \begin{tabular}{lrrrr}
        \toprule
        \textbf{Ablation} & \makecell{\textbf{Drug}\\\textbf{offences}} & \makecell{\textbf{Medical}\\\textbf{errors}} & \makecell{\textbf{Personal}\\\textbf{rights}} & \textbf{Overall} \\
        \midrule
        Full (\system{}) & \phantom{+}83.9\% & \phantom{+}51.5\% & \phantom{+}64.0\% & \phantom{+}66.5\%$\pm$16.3pp \\
        w/o PDH & -31.5pp & \phantom{0}+5.5pp & -16.9pp & -14.3pp$\pm$18.7pp \\
        w/o CBR & \phantom{0}+0.1pp & \phantom{0}+0.2pp & \phantom{0}+4.4pp & \phantom{0}+1.5pp$\pm$\phantom{0}2.5pp \\
        w/o DGR & \phantom{0}+0.7pp & \phantom{0}+3.9pp & \phantom{0}-7.4pp & \phantom{0}-1.0pp$\pm$\phantom{0}5.8pp \\
        w/o PDH+CBR & -36.2pp & \phantom{0}-7.4pp & \phantom{0}-4.3pp & -16.0pp$\pm$17.5pp \\
        w/o PDH+DGR & -29.4pp & -13.7pp & -18.8pp & -20.6pp$\pm$\phantom{0}8.0pp \\
        w/o CBR+DGR & -40.0pp & \phantom{0}-6.2pp & \phantom{0}-9.6pp & -18.6pp$\pm$18.6pp \\
        \makecell[l]{w/o PDH+CBR\\+DGR} & -83.9pp & -33.1pp & -43.3pp & -53.4pp$\pm$26.9pp \\
        \bottomrule
        \end{tabular}%
    }
    \caption{Expert-question coverage after selectively ablating components of the \system{} pipeline. Entries show the change in coverage (percentage points, pp) relative to the full system, averaged over the research problems.}
    \label{tab:ablation}
\end{table}

\section{Costs}
\label{sec:appendix:evaluation:costs}
\begin{table}[!htb]
    \centering
    \small
    \setlength{\tabcolsep}{4.5pt}
    \resizebox{\linewidth}{!}{%
    \begin{tabular}{lccc}
        \toprule
        \textbf{Backbone} & \textbf{vanilla LLM} & \textbf{ScheMatiQ} & \textbf{\system{}} \\
        \midrule
        gpt-5.4 & \$0.024 $\pm$ 0.008 & \$0.176 $\pm$ 0.076 & \$3.785 $\pm$ 0.312 \\
        gpt-5.4-mini & \$0.005 $\pm$ 0.000 & \$0.044 $\pm$ 0.020 & \$0.828 $\pm$ 0.059 \\
        gpt-5.4-nano & \$0.002 $\pm$ 0.001 & \$0.014 $\pm$ 0.007 & \$0.234 $\pm$ 0.015 \\
        claude-sonnet-4.6 & \$0.027 $\pm$ 0.006 & \$0.243 $\pm$ 0.086 & \$3.057 $\pm$ 0.510 \\
        qwen3.6-35b-a3b & \$0.005 $\pm$ 0.001 & \$0.033 $\pm$ 0.007 & \$0.183 $\pm$ 0.010 \\
        gemma-4-e4b-it & \$0.000 $\pm$ 0.000 & \$0.016 $\pm$ 0.001 & \$0.041 $\pm$ 0.001 \\
        llama-4-scout-17b & \$0.000 $\pm$ 0.000 & \$0.007 $\pm$ 0.001 & \$0.023 $\pm$ 0.002 \\
        \bottomrule
        \end{tabular}%
    }
    \caption{The LLM inference cost in USD per research problem ($mean \pm std$), per backbone, for the \emph{vanilla LLM}, \emph{ScheMatiQ}, and \system{} configurations.}
    \label{tab:costs}
\end{table}

\Cref{tab:costs} reports the average inference cost per research problem for each configuration. \System{} is consistently the most expensive configuration, reflecting its multi-agent, multi-round design, while ScheMatiQ costs roughly an order of magnitude less than \system{} and vanilla LLM is cheapest by a further order of magnitude. This gap is most pronounced for large closed-weight backbones (e.g.,~gpt-5.4, claude-sonnet-4.6) and narrows substantially for smaller open-weight models (e.g.,~gemma-4-e4b-it, llama-4-scout-17b), where \system{}'s absolute cost drops to a few cents per problem. Combined with the coverage results in \Cref{tab:results}, this suggests that \system{}'s largest coverage gains come at a real cost premium, but that premium can be kept small by pairing \system{} with a smaller backbone rather than the largest available model.

\end{document}